\documentclass[runningheads]{llncs}

\usepackage{amsmath}
\usepackage{amssymb}
\usepackage{esvect}
\usepackage[table]{xcolor}   
\definecolor{cprred}{RGB}{255,230,230} 
\usepackage[T1]{fontenc}
\usepackage{graphicx,verbatim}
\usepackage{xcolor}
\usepackage{makecell}
\usepackage{booktabs}
\usepackage{adjustbox}
\usepackage{array}
\usepackage{overpic}
\usepackage{multirow}
\usepackage[colorlinks=true,
            linkcolor=blue,
            citecolor=blue,
            urlcolor=blue]{hyperref}
\usepackage{bbding} 
\begin{document}
\setlength{\abovedisplayskip}{3pt}
\setlength{\belowdisplayskip}{3pt}
\setlength{\abovedisplayshortskip}{3pt}
\setlength{\belowdisplayshortskip}{3pt}
\setlength{\dbltextfloatsep}{6pt}  

\setlength{\textfloatsep}{6pt}     
\setlength{\abovecaptionskip}{2pt}
\setlength{\belowcaptionskip}{0pt}
\title{\textbf{CPR}: \underline{\textbf{C}}hained \underline{\textbf{P}}erceptual \underline{\textbf{R}}efinement for Coarse-to-Fine Medical Image Classification}
\titlerunning{CPR for Medical Imaging Classification}
%

\author{
Si-Yuan Lu\inst{1,*}
\and Hanruo Zhu\inst{2,*}
\and Ziquan Zhu\inst{2,*}
\and Gaojie Jin\inst{3}
\and Zeyu Fu\inst{3}
\and Lu Yin\inst{4}(\Envelope)
\and Ke Li\inst{3}
\and Lu Liu\inst{3}
\and Tianjin Huang\inst{3}(\Envelope)
}

\authorrunning{S.-Y. Lu et al.}

\institute{
Nanjing University of Posts and Telecommunications, Nanjing, China
\and
University of Leicester, Leicester, UK
\and
University of Exeter, Exeter, UK
\and
University of Surrey, Guildford, UK\\
\email{t.huang2@exeter.ac.uk;l.yin@surrey.ac.uk}\\
\textsuperscript{*}Equal contribution.
}

  
\maketitle              
\begin{abstract}
High-resolution medical images contain fine-grained, spatially sparse cues that are critical for diagnosis, yet preserving full resolution incurs substantial computational and memory costs. Most deep models process images uniformly, leading to redundant computation or loss of diagnostic detail under downsampling. We propose \underline{C}hained \underline{P}erceptual \underline{R}efinement (\texttt{CPR}), a coarse-to-fine framework that formulates medical image analysis as a sequential global-to-local decision process. Starting from a low-resolution global view, \texttt{CPR} dynamically predicts the location and spatial extent of refinement regions, extracts high-resolution evidence from the original image, and incrementally integrates it with global context. By keeping the backbone input size fixed while contracting the perceptual field, \texttt{CPR} preserves diagnostic fidelity with constant peak GPU memory. Extensive experiments on five medical imaging datasets and multiple backbone architectures demonstrate that \texttt{CPR} consistently outperforms both fixed-resolution and multi-scale state-of-the-art baselines, achieving improvements of up to 2.27\% over the second-best method. It also achieves up to a 19.6$\times$ reduction in GFLOPs at matched accuracy, establishing a superior accuracy–efficiency trade-off for high-resolution medical image analysis. The code is available at: \href{https://github.com/SiyuanLuLSY/CPR-Chained-Perceptual-Refinement}{GitHub}.


\end{abstract}

\section{Introduction}

High-resolution medical images are central to modern clinical workflows, as diagnostically decisive cues are often fine-grained and spatially sparse. Across modalities such as digital pathology, high-resolution radiography, and CT, clinically meaningful evidence may manifest as subtle textural variations, small morphological structures, or tiny lesions occupying only limited regions\cite{yu_siamese-transport_2024,wang_joint_2024}. Consequently, many medical tasks including detection, segmentation, and treatment planning, must jointly leverage global anatomical context and local high-frequency details~\cite{sabottke2020effect,haque2023effect,ashman2022whole}.  \emph{This creates a fundamental trade-off}: preserving diagnostic resolution incurs substantial memory and computation. As shown in Fig.~\ref{fig:fig1}(a), increasing input resolution generally improves accuracy, yet GPU memory consumption rises sharply and quickly becomes impractical. Conversely, aggressive downsampling reduces cost at the expense of diagnostic fidelity (Fig.~\ref{fig:fig1}(b)) , exposing the limitation of fixed-resolution paradigms that uniformly allocate computation across spatially heterogeneous data.

While dominant deep learning architectures have achieved remarkable success, they predominantly rely on spatially uniform computation \cite{murugesan_neural_2024}. Standard approaches process inputs at fixed dimensions or utilize dense patch-based extraction \cite{vats_incremental_2024,serrao_automatic_2024}, uniformly allocating compute across an image regardless of regional informativeness~\cite{kaiming_he_deep_2016,alexey_dosovitskiy_image_2021}. Multi-scale architectures (e.g., pyramidal features) incorporate multiple resolutions, yet their computation remains predefined and often redundant~\cite{swinvit,ryali2023hiera,ZoomNeXt,mgca}. More importantly, these paradigms are passive: models process inputs with fixed resolution and static allocation, without adaptively deciding \emph{where} and \emph{at what scale} to focus. \emph{In contrast}, human perception is sequential and selective. Clinicians rarely analyze medical images in a single pass; instead, they form a global impression and progressively refine attention toward diagnostically relevant regions, integrating global context with localized evidence. While recent advancements such as AdaptiveNN~\cite{wang2025emulating,meng_lou_overlock_2025} explore active, human-inspired vision mechanisms by explicitly modeling sequential fixations, they predominantly rely on extracting predefined, fixed-size crops. In medical image analysis, where subtle diagnostic evidence is strictly anchored to its broader anatomical context, such unconstrained, disjointed patch sampling risks redundant observations and disrupts the spatial hierarchy.

Motivated by foveal vision, we introduce \underline{C}hained \underline{P}erceptual \underline{R}efinement (\texttt{CPR}), a perception-inspired framework that formulates medical image analysis as a dynamic, coarse-to-fine sequential decision process. Starting from a low-resolution global view, \texttt{CPR} iteratively allocates a localized \emph{perceptual field} by dynamically predicting refinement parameters, specifically, both top-left coordinates and spatial extent, at each step. The selected regions are extracted from the original high-resolution image and incrementally fused with the global context to update the model's prediction. Unlike Adaptive NN, which mainly uses fixed-size local observations, \texttt{CPR} jointly adapts crop location and scale; unlike OverLoCK, it explicitly revisits the original high-resolution image rather than relying on feature-level context mixing; and unlike PAMIL, it targets image-level classification instead of WSI-level MIL. We summarize our main contributions as follows:

\begin{figure}[!t]
\centering

\begin{minipage}[t]{0.6\linewidth}
  \centering
  \includegraphics[width=\linewidth,height=0.16\textheight,keepaspectratio]{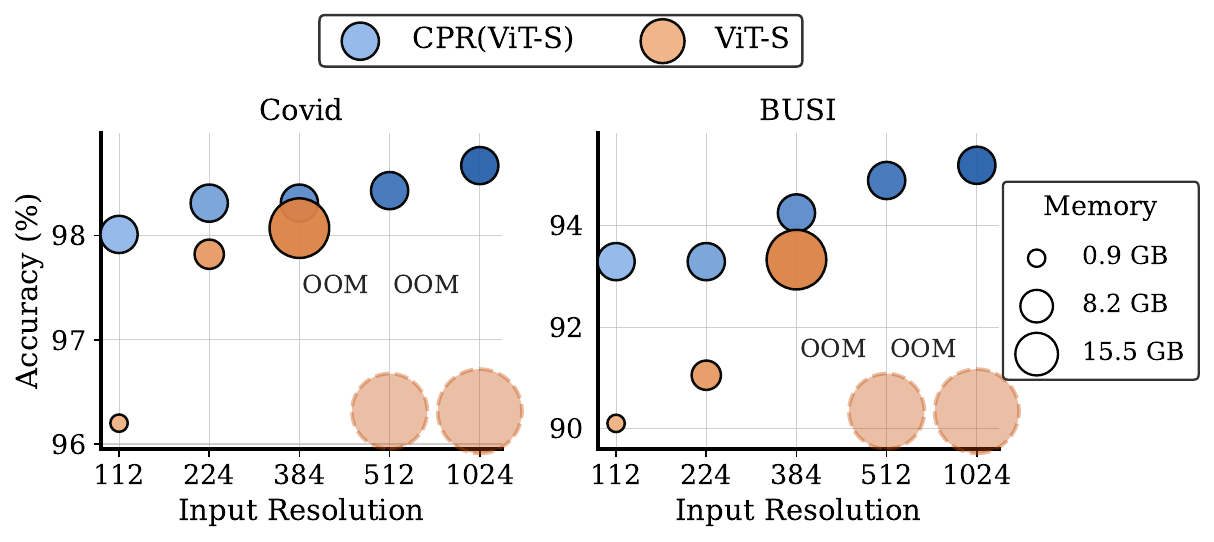}\\[-0.6em]
  {\small (a)}
\end{minipage}%
\hspace{0.0001\linewidth}%
\begin{minipage}[t]{0.392\linewidth}
  \centering
  \includegraphics[width=\linewidth,height=0.16\textheight,keepaspectratio,
    trim=0.1cm 0.5cm 0.5cm 0.5cm,clip]{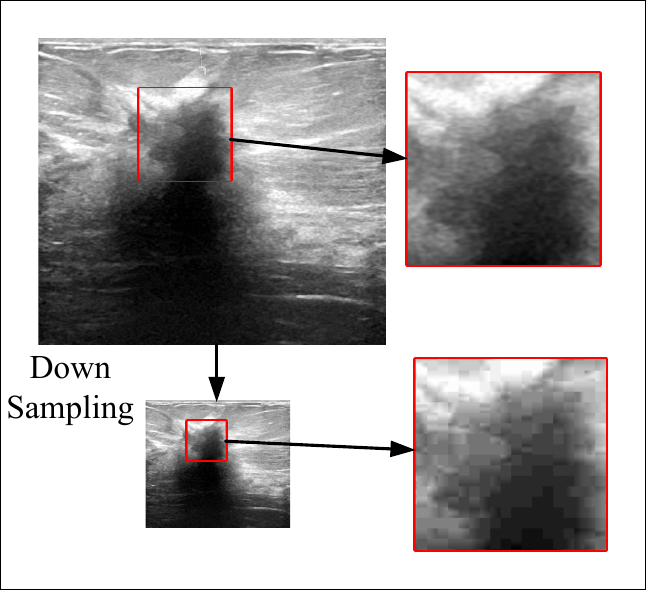}\\[-0.6em]
  {\small (b)}
\end{minipage}

\caption{
\textbf{Resolution--efficiency trade-off in medical image analysis.}
(a) Accuracy and peak GPU memory at different input resolutions on COVID and BUSI (RTX 3090, batch size 64). Bubble size denotes memory; OOM indicates infeasible settings.
(b) Downsampling removes fine-grained details.
}
\label{fig:fig1}

\end{figure}

\begin{itemize}
\item[$\star$] \textbf{Method.} We propose \texttt{CPR}, a framework that transitions medical image analysis from static processing to an active, sequential paradigm. By explicitly modeling global-to-local transitions through dynamic scale contraction, \texttt{CPR} progressively restricts its perceptual field to diagnostically relevant regions, emulating the clinical coarse-to-fine diagnostic workflow.

\item[$\star$] \textbf{SOTA Performance.} \texttt{CPR} demonstrates superior performance across four modalities and multiple architectures. Specifically, it consistently surpasses the current state-of-the-art (SOTA) baseline, Adaptive NN~\cite{wang2025emulating}, across the Shenzhen, Covid, BUSI, BreaKHis, and EyePACs datasets, with improvements of 2.27\%, 0.24\%, 1.92\%, 0.45\%, and 1.42\%, respectively. 

\item[$\star$] \textbf{Memory and Computational Efficiency.}  By employing a selective refinement strategy with a fixed backbone input size, \texttt{CPR} ensures constant peak GPU memory regardless of original image resolution, effectively bypassing quadratic memory scaling. This framework establishes a superior accuracy–efficiency trade-off, delivering up to a 19.6$\times$ reduction in GFLOPs compared to dense high-resolution processing at equivalent accuracy levels.
\end{itemize}

\section{Chained Perceptual Refinement Framework}

\begin{figure}[tbh]
    \centering
    \includegraphics[trim=0.5cm 0.5cm 0.5cm 1.9cm, clip, width=0.86\linewidth]{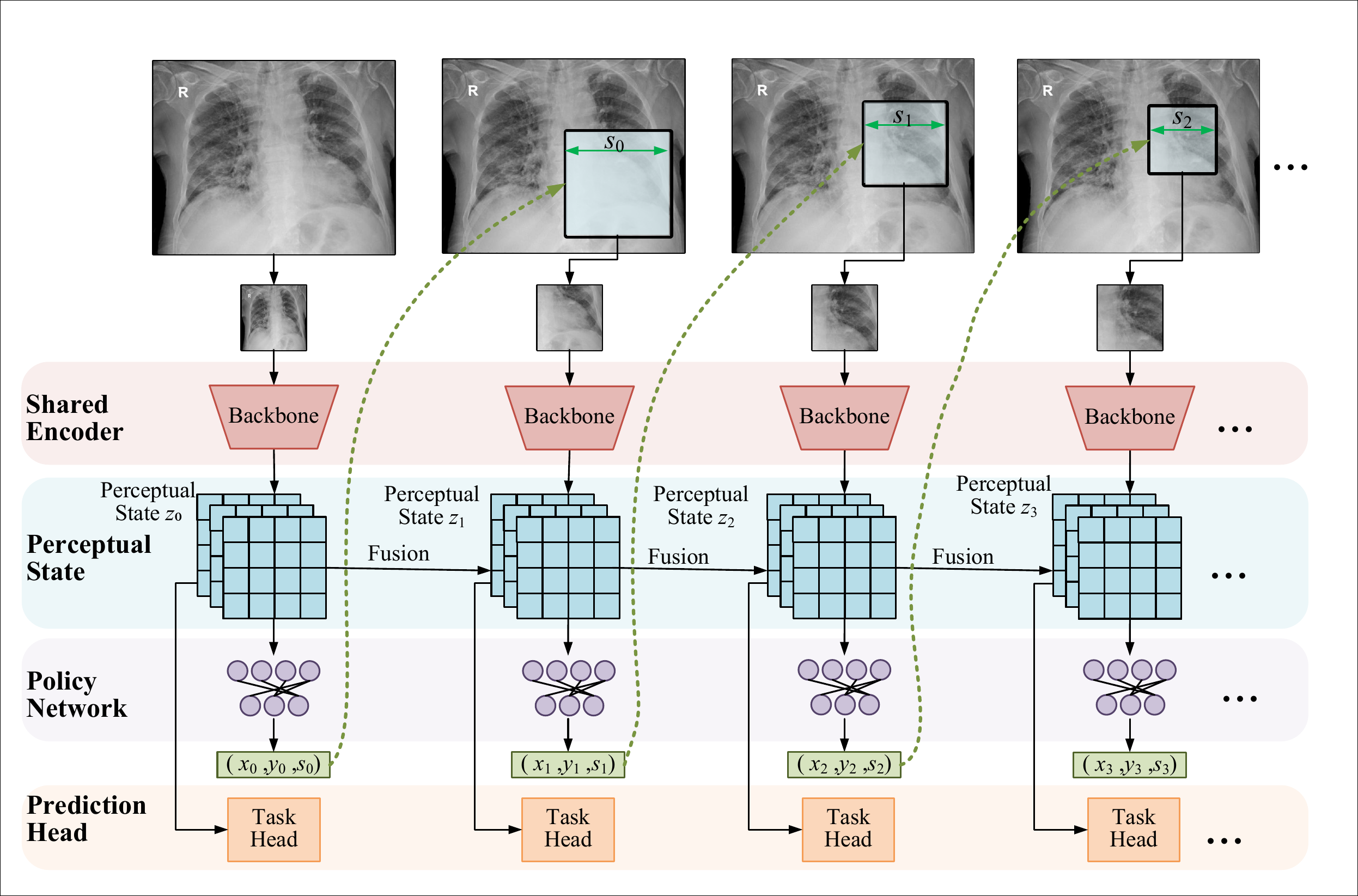}
    \caption{\textbf{Overview of the \texttt{CPR} framework.} Inspired by human coarse-to-fine visual perception, \texttt{CPR} models medical image interpretation as a sequential refinement process that progressively focuses on localized regions while maintaining constant input resolution, enabling efficient and accurate analysis.}
    \label{fig:cpr}

\end{figure}

We propose Chained Perceptual Refinement (\texttt{CPR}), a sequential coarse-to-fine framework that dynamically allocates perceptual fields while maintaining constant computational cost (Fig.~\ref{fig:cpr}). 
The framework consists of four core modules: a \emph{Shared Encoder}, a \emph{Perceptual State} update mechanism, a \emph{Policy Network} for refinement, and a \emph{Prediction Head} for task supervision. 
Instead of uniformly processing the entire image at high resolution, \texttt{CPR} formulates perceptual refinement as an instance-adaptive decision process over spatial scale and location. 
This design enables explicit and controllable compute allocation, allowing the model to progressively focus on diagnostically informative regions without increasing input resolution.

\noindent\textbf{Shared Encoder.} Given an input image $X \in \mathbb{R}^{H \times W \times C}$, \texttt{CPR} first constructs a global observation by resizing the image to a canonical resolution. 
The resized image is processed by a shared backbone $f_{\theta}$, which is reused across all refinement steps. 
Unlike conventional multi-scale strategies that increase input resolution, \texttt{CPR} keeps the input size fixed and instead adapts the perceptual field over the original image, decoupling perceptual granularity from computational cost. 
This design enables refinement to be achieved through spatial selection rather than resolution scaling, avoiding the quadratic memory growth associated with high-resolution processing. The initial perceptual representation is defined as
\begin{equation}
\mathbf{z}_0 = f_{\theta}(X_{\mathrm{global}}).
\end{equation}


\noindent\textbf{Perceptual State.} \texttt{CPR} maintains a latent \emph{perceptual state} that aggregates information across refinement steps, serving as an evolving representation of accumulated evidence. 
At step $t$, the localized observation $X_t$ is encoded as
\begin{equation}
\mathbf{Z}_t = f_{\theta}(X_t).
\end{equation}
\noindent The perceptual state is updated via residual fusion: $\mathbf{H}_{t+1} = \mathbf{H}_t + \mathbf{W}_t \mathbf{Z}_t$ where $\mathbf{H}_t$ denotes the accumulated state and $\mathbf{W}_t$ is a learned projection matrix. 
This update progressively integrates the global context with localized evidence without increasing input resolution, forming a sequential refinement trajectory. 
Importantly, the state transition $\mathbf{H}_t \rightarrow \mathbf{H}_{t+1}$ encodes the dependency between successive perceptual decisions, allowing future refinement steps to be conditioned on accumulated evidence.

\noindent\textbf{Policy Network.} The refinement process is governed by a policy network $g_{\phi}$, which predicts the parameters of the perceptual field conditioned on the current state:
\begin{equation}
\mathbf{a}_t = (x_t, y_t, s_t) \sim g_{\phi}(\mathbf{H}_t),
\end{equation}
where $(x_t, y_t)$ denotes the normalized top-left coordinate and $s_t$ denotes the normalized side length of the cropped region. The corresponding crop is extracted from the original image with top-left position $(x_t W, y_t H)$ and spatial size $(s_t W, s_t H)$, and is then resized to the canonical input size before being processed by the shared encoder. These parameters determine \emph{where} to attend and \emph{at what scale} to allocate computation. Throughout the refinement stages, the perceptual scales are encouraged to contract, forming a structured coarse-to-fine trajectory. By conditioning on $\mathbf{H}_t$, the policy learns an instance-adaptive refinement strategy rather than a fixed spatial schedule.

\noindent\textbf{Prediction Head.} At each refinement stage, a task-specific prediction head operates on the updated perceptual state:
\begin{equation}
\mathbf{o}_t = h(\mathbf{H}_t),
\end{equation}
where $h(\cdot)$ denotes the task head.

\texttt{CPR} produces a global prediction $\mathbf{o}^{\mathrm{glob}}$ and refined local predictions $\{\mathbf{o}^{\mathrm{loc}}_t\}_{t=1}^{T}$, 
allowing progressive error correction across refinement steps and enabling intermediate supervision to stabilize training.

\section{Training Objective}
\noindent\textbf{Supervised prediction learning.}
The perceptual backbone (global and local prediction heads) is trained with standard task supervision. 
Classification loss is applied to the global prediction and all localized predictions:
\begin{equation}
\mathcal{L}_{\mathrm{pred}}
=
\ell(\mathbf{o}^{\mathrm{glob}}, y)
+
\sum_{t=1}^{T} \ell(\mathbf{o}^{\mathrm{loc}}_{t}, y).
\end{equation}

\noindent\textbf{Prediction consistency.}
To stabilize progressive refinement, the final localized prediction 
$\mathbf{o}^{\mathrm{loc}}_{T}$ is treated as a soft reference. 
Intermediate predictions are aligned with it via KL divergence:
\begin{equation}
\mathcal{L}_{\mathrm{sup}}
=
\mathcal{L}_{\mathrm{pred}}
+
\alpha \mathcal{L}_{\mathrm{cons}},\;\;
\mathcal{L}_{\mathrm{cons}}
=
\mathrm{KL}(\mathbf{o}^{\mathrm{glob}} \,\|\, \mathbf{o}^{\mathrm{loc}}_{T})
+
\sum_{t=1}^{T-1}
\mathrm{KL}(\mathbf{o}^{\mathrm{loc}}_{t} \,\|\, \mathbf{o}^{\mathrm{loc}}_{T}).
\end{equation}


\noindent\textbf{Refinement regularization.}
To encourage structured coarse-to-fine behavior, we introduce two reward regularizers:
\begin{equation}
\phi^{\mathrm{scale}}_{t} = \max(0, s_t - s_{t-1}), 
\quad
\phi^{\mathrm{rep}}_{t} = \mathbb{I}[\hat{\mathbf{a}}_{t} = \hat{\mathbf{a}}_{t-1}],
\end{equation}
where $s_t$ denotes the perceptual scale and $\hat{\mathbf{a}}_t$ is the discretized refinement action. 
The scale regularizer penalizes non-decreasing perceptual scales, while the repetition regularizer discourages consecutive identical refinement actions. 
The final shaped reward is defined as
$
\tilde{r}_t
=
r_t
-
\lambda_{\mathrm{scale}} \phi^{\mathrm{scale}}_{t}
-
\lambda_{\mathrm{rep}} \phi^{\mathrm{rep}}_{t}.
$

\noindent\textbf{Progressive refinement as reinforcement learning.}
The refinement parameters are generated by a policy network 
$g_{\phi}(\mathbf{a}_t \mid \mathbf{H}_t)$, 
where $\mathbf{H}_t$ denotes the accumulated perceptual state and 
$\mathbf{a}_t$ encodes the region extraction parameters at step $t$.
The policy is optimized using Proximal Policy Optimization (PPO), treating perceptual refinement as a sequential decision-making problem over spatial scale and location. 
We define a step-wise refinement reward based on the reduction in task loss:
$
r_t
=
\ell(\mathbf{o}_{t}, y)
-
\ell(\mathbf{o}_{t+1}, y)
$
where $\mathbf{o}_0 = \mathbf{o}^{\mathrm{glob}}$ and 
$\mathbf{o}_t = \mathbf{o}^{\mathrm{loc}}_t$ for $t \geq 1$.

\section{Experiments}

\subsection{Experiment Setup}

\noindent\textbf{Datasets and Baselines.}
We evaluate \texttt{CPR} on five publicly available medical imaging benchmarks
:
(1) \textbf{Shenzhen Chest X-ray} \cite{jaeger_s_automatic_2014},
(2) \textbf{COVID-19 Radiography Database} \cite{muhammad_e_h_chowdhury_can_2020},
(3) \textbf{BUSI Breast Ultrasound} \cite{al-dhabyani_dataset_2020},
(4) \textbf{BreaKHis 400$\times$} \cite{breakhis},
and (5) \textbf{EyePACS-AIROGS} \cite{eyepac}.
These datasets span X-ray, ultrasound, histopathology, and retinal fundus photography,
with substantial variation in native resolution. We compare \texttt{CPR} with convolutional and transformer backbones (ResNet-50, ViT-S, Swin-S) under identical training settings. We further evaluate against recent medical imaging methods, including feature fusion (Adaptive NN, OverLock, Hiera, GLNet), feature alignment (mammo-CLIP, MGCA), and multi-instance learning (GABMIL, PAMIL).

\noindent\textbf{Implementation Details.}
Models are trained for 100 epochs using AdamW with batch size 64. \texttt{CPR} and multi-instance learning methods operate on dataset-adaptive high resolutions (up to $2048^2$, $1024^2$, and $512^2$), resized according to native image scales, while other baseline models use fixed input size.

\begin{table}[tbh]
\centering
\caption{
Comparison between \texttt{CPR} and vanilla backbones using ResNet-50, ViT-Small and Swin-Small.
NA: no valid result due to OOM; OOM: out-of-memory.
}
\label{tab:cpr_backbone_comparison}
\small
\setlength{\tabcolsep}{2pt}
\resizebox{0.78\linewidth}{!}{
\begin{tabular}{lccc ccc ccc}
\toprule
\multirow{2}{*}{Dataset}
& \multicolumn{3}{c}{ResNet-50}
& \multicolumn{3}{c}{ViT-S}
& \multicolumn{3}{c}{Swin-S} \\
\cmidrule(lr){2-4}\cmidrule(lr){5-7}\cmidrule(lr){8-10}
& (224) & (384) & \cellcolor{cprred}\texttt{CPR}
& (224) & (384) & \cellcolor{cprred}\texttt{CPR}
& (224) & (448) & \cellcolor{cprred}\texttt{CPR} \\
\midrule
Shenzhen  & 87.88 & 89.39 & \cellcolor{cprred}\textbf{90.15} & 76.52 & 86.36 & \cellcolor{cprred}\textbf{90.15} & 88.64 & NA & \cellcolor{cprred}\textbf{90.91} \\
COVID-19  & 97.10 & 97.28 & \cellcolor{cprred}\textbf{98.25} & 93.30 & 98.07 & \cellcolor{cprred}\textbf{98.67} & 96.20 & NA & \cellcolor{cprred}\textbf{98.43} \\
BUSI      & 91.05 & 91.10 & \cellcolor{cprred}\textbf{92.01} & 89.14 & 93.33  & \cellcolor{cprred}\textbf{94.89} & 90.73 & NA & \cellcolor{cprred}\textbf{92.97} \\
BreaKHis  & 94.31 & 96.70 & \cellcolor{cprred}\textbf{98.35} & 97.06 & 98.53 & \cellcolor{cprred}\textbf{99.27} & 96.88 & NA & \cellcolor{cprred}\textbf{98.53} \\
EyePACS    & 94.16 & 94.68 & \cellcolor{cprred}\textbf{94.94} & 94.03 & 94.42 & \cellcolor{cprred}\textbf{95.19} & 94.16 & NA & \cellcolor{cprred}\textbf{94.81} \\
\midrule
Memory (G) & 3.1 & 12.1 & 5.7 & 3.5 & 15.8 & 8.1 & 8.3 & OOM & 12.7 \\
\bottomrule
\end{tabular}}
\end{table}

\begin{table}[t]
\centering
\caption{Classification accuracy (\%) and memory consumption across five public datasets. The best results are in \textbf{bold}, and the second-best are \underline{underlined}.}
\label{tab:overall_results}
\small
\setlength{\tabcolsep}{4pt}
\renewcommand{\arraystretch}{1.0}

\resizebox{\linewidth}{!}{
\setlength{\tabcolsep}{4pt}
\renewcommand{\arraystretch}{1.0}
\begin{tabular}{@{}
p{2.6cm}
>{\centering\arraybackslash}m{1.95cm}
>{\centering\arraybackslash}m{1.55cm}
>{\centering\arraybackslash}m{1.55cm}
>{\centering\arraybackslash}m{1.35cm}
>{\centering\arraybackslash}m{1.55cm}
>{\centering\arraybackslash}m{1.35cm}
c
@{}}
\hline
\textbf{} & \textbf{Venue} & \textbf{Shenzhen} & \textbf{Covid} & \textbf{BUSI} & \textbf{BreaKHis} & \textbf{EyePACS} & \textbf{Mem. (G)} \\
\hline
Mean Resolution & -- & 2699$\times$2794 & 1302$\times$977 & 615$\times$501 & 700$\times$459 & 512$\times$512 & -- \\
\hline
Adaptive NN \cite{wang2025emulating} & NMI'2025   & 87.88 & \underline{98.43} & \underline{92.97} & \underline{98.72} & \underline{93.77} & \textbf{8.1} \\
OverLoCK \cite{meng_lou_overlock_2025} & CVPR'2025 & \underline{89.39} & 97.89 & \underline{92.97} & 97.98 & 94.03 & 23.8 \\
Hiera \cite{ryali2023hiera} & ICML'2023       & 85.61 & 98.19 & 91.37 & 93.76 & 94.81 & \textbf{8.1} \\
GLNet \cite{zhu2024glnet} & NIPS'2024         & 89.39 & 98.43 & 92.33 & 98.53 & 94.16 & 12.4 \\
mammo-CLIP \cite{mammo-clip} & MICCAI'2024     & 88.64 & 97.52 & 85.62 & 86.97 & 94.29 & 10.5 \\
MGCA \cite{mgca} & EAAI'2025                  & \underline{89.39} & 97.89 & 92.01 & 97.43 & 94.42 & 8.3 \\
GABMIL \cite{gbmil2025} & MICCAI'2025          & 86.36 & 96.98 & 84.66 & 91.19 & 94.42 & 8.3 \\
PAMIL \cite{pamil2024} & CVPR'2024             & 84.85 & 94.63 & 83.07 & 90.64 & 95.06 & 11.3 \\
\rowcolor{cprred} \texttt{CPR} & --            & \textbf{90.15} & \textbf{98.67} & \textbf{94.89} & \textbf{99.27} & \textbf{95.19} & \textbf{8.1} \\
\hline
\multicolumn{8}{@{}l@{}}{\footnotesize \textit{Note:} NMI denotes \textit{Nature Machine Intelligence}; EAAI denotes \textit{Engineering Applications of Artificial Intelligence}.} \\
\end{tabular}
}
\end{table}

\subsection{Main Results}

\noindent\textbf{Effectiveness of \texttt{CPR}.}
Table~\ref{tab:cpr_backbone_comparison} compares \texttt{CPR} with ResNet-50, ViT-S, and Swin-S. \texttt{CPR} consistently outperforms fixed-resolution baselines, showing that gains stem from coarse-to-fine refinement rather than backbone choice. Naive resolution scaling (ResNet-50/ViT-S at $384^2$) provides limited benefit. Due to its window-based architecture requiring strict resolution divisibility, Swin-S cannot adopt $384^2$ and instead uses $448^2$, which results in OOM. In contrast, \texttt{CPR} achieves higher accuracy with lower memory consumption; for example, it improves accuracy by 3.79\% while reducing peak GPU memory by 48.73\% compared to ViT-S on Shenzhen.

\noindent\textbf{State-of-the-Art Comparisons}. Table~\ref{tab:overall_results} provides a comprehensive evaluation of \texttt{CPR} against competitive baselines across five public datasets. \texttt{CPR} attains the highest classification accuracy on all benchmarks, consistently surpassing established methods such as Adaptive NN and OverLoCK. Specifically, \texttt{CPR} outperforms the second-best method, Adaptive NN, by margins of 2.27\%, 0.24\%, 1.92\%, 0.45\%, and 1.42\% on the Shenzhen, Covid, BUSI, BreaKHis, and EyePACs datasets, respectively. Furthermore, \texttt{CPR} maintains a minimal memory footprint of 8.1 G, matching the most efficient baselines while processing images at their native, dataset-adaptive resolutions. 


\begin{figure*}[tbh]
    \centering
    \includegraphics[width=\textwidth]{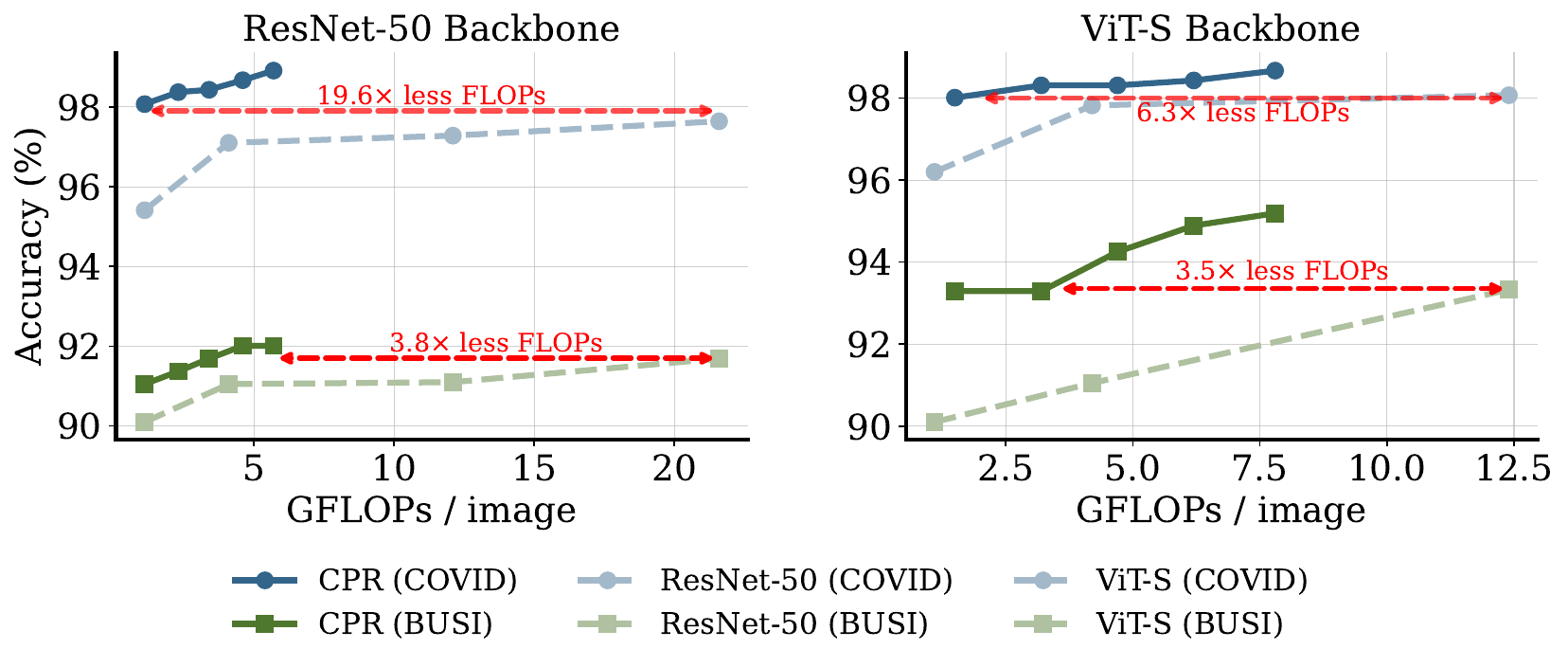}
    \caption{
Accuracy–efficiency trade-offs of CPR (solid) versus fixed-resolution baselines (dashed) on COVID-19 and BUSI.
Baselines use resolutions $\{112,224,384,512\}$ for ResNet-50 and $\{112,224,384\}$ for ViT-S (higher settings may exceed memory).
Red dashed arrows denote GFLOPs reduction at comparable accuracy.
}
    \label{fig:tradeoff_covid_busi}
\end{figure*}

\begin{figure*}[tbh]
\centering
\setlength{\tabcolsep}{0pt} 
\renewcommand{\arraystretch}{0.9}

\begin{tabular}{@{}>{\centering\arraybackslash}m{0.19\textwidth}@{\hspace{3pt}}
                >{\centering\arraybackslash}m{0.19\textwidth}@{\hspace{3pt}}
                >{\centering\arraybackslash}m{0.19\textwidth}@{\hspace{3pt}}
                >{\centering\arraybackslash}m{0.19\textwidth}@{\hspace{3pt}}
                >{\centering\arraybackslash}m{0.19\textwidth}@{}}
\textbf{Shenzhen} & \textbf{Covid} & \textbf{BUSI} & \textbf{BreaKHis} & \textbf{EyePACS} \\

\adjustbox{max width=\linewidth, max height=2.7cm}{\includegraphics{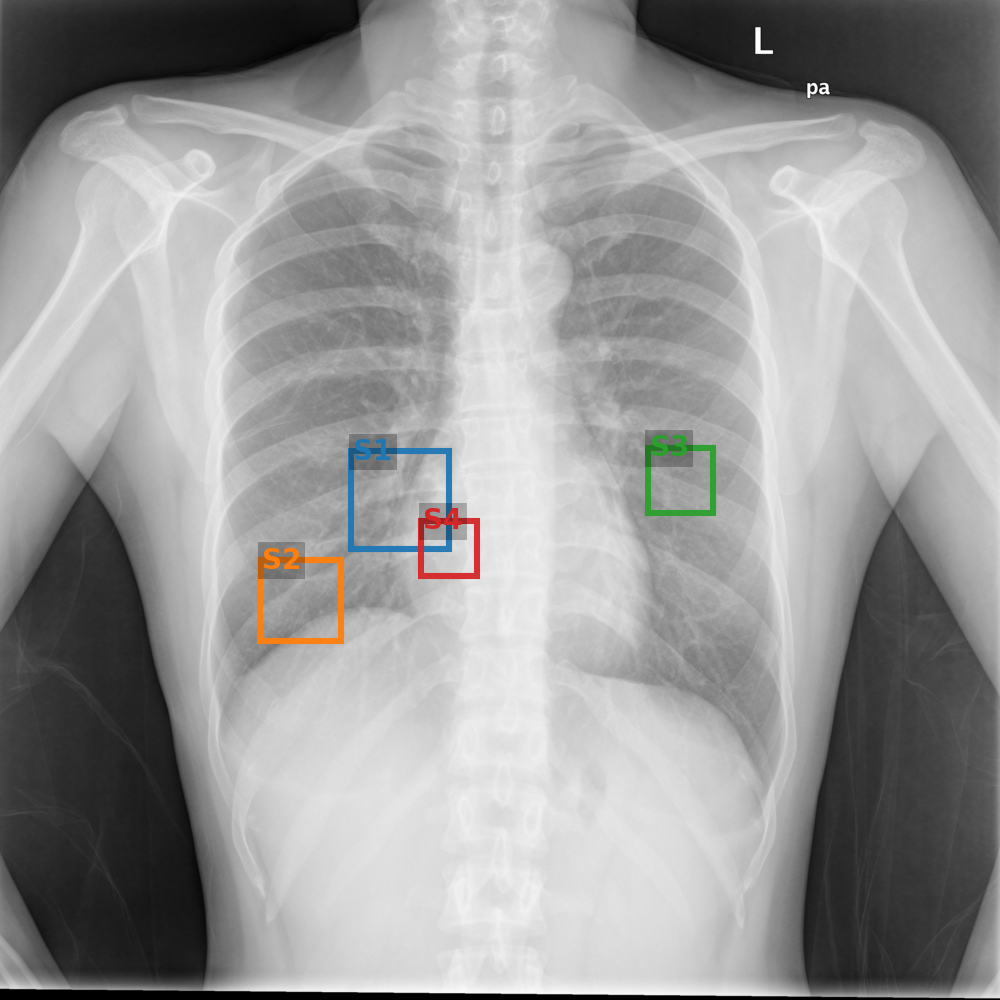}} &
\adjustbox{max width=\linewidth, max height=2.7cm}{\includegraphics{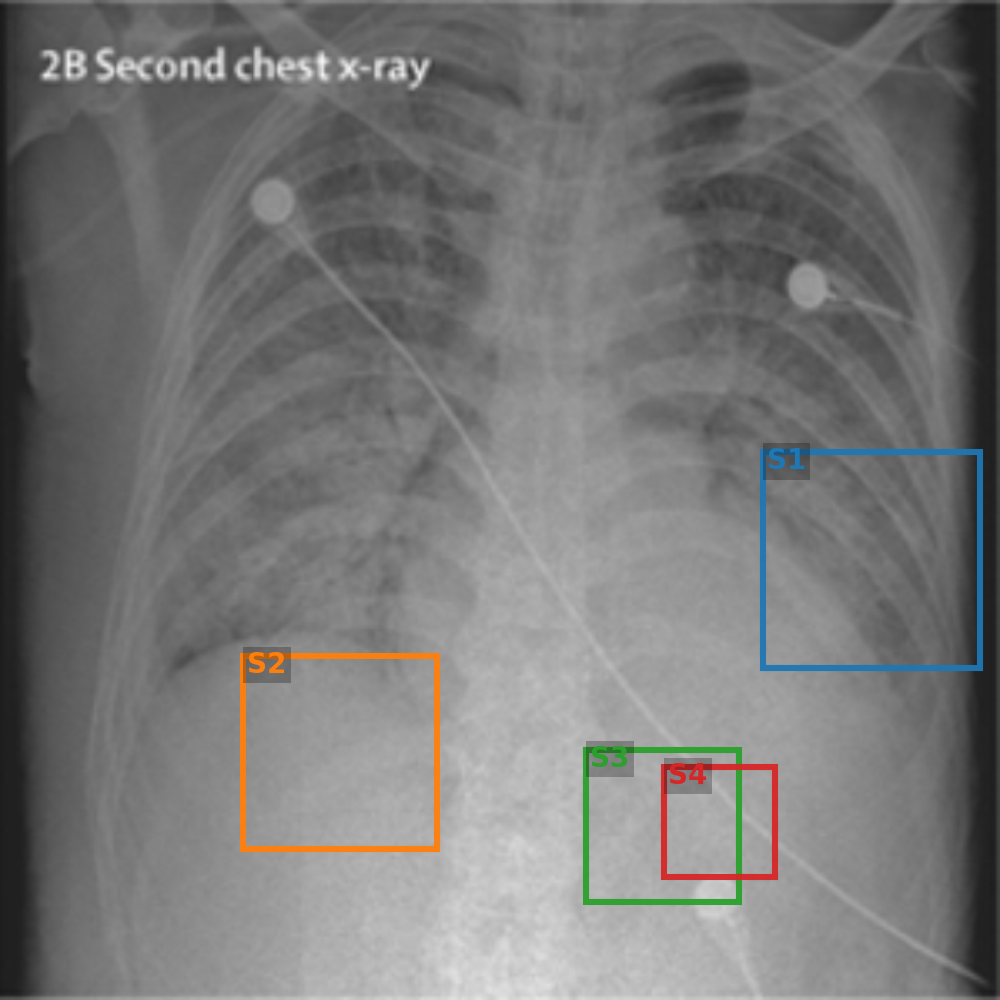}} &
\adjustbox{max width=\linewidth, max height=2.7cm}{\includegraphics{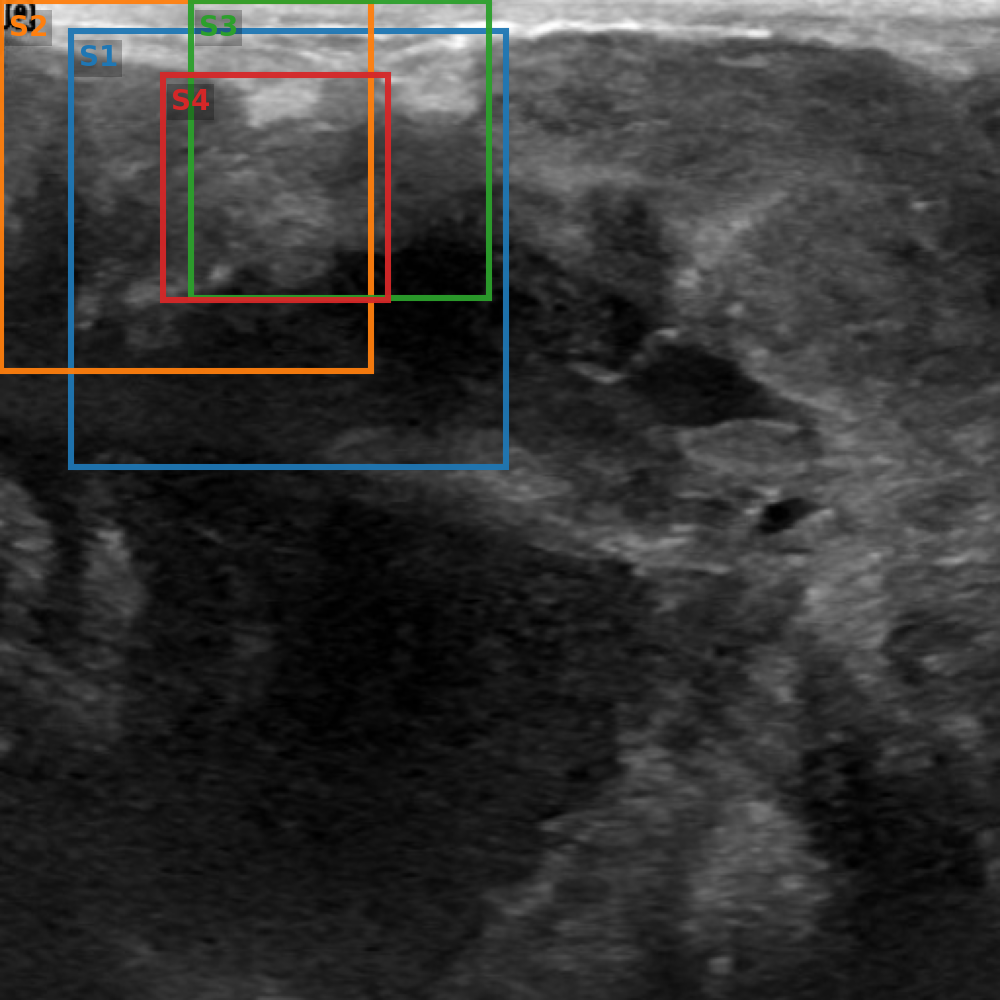}} &
\adjustbox{max width=\linewidth, max height=2.7cm}{\includegraphics{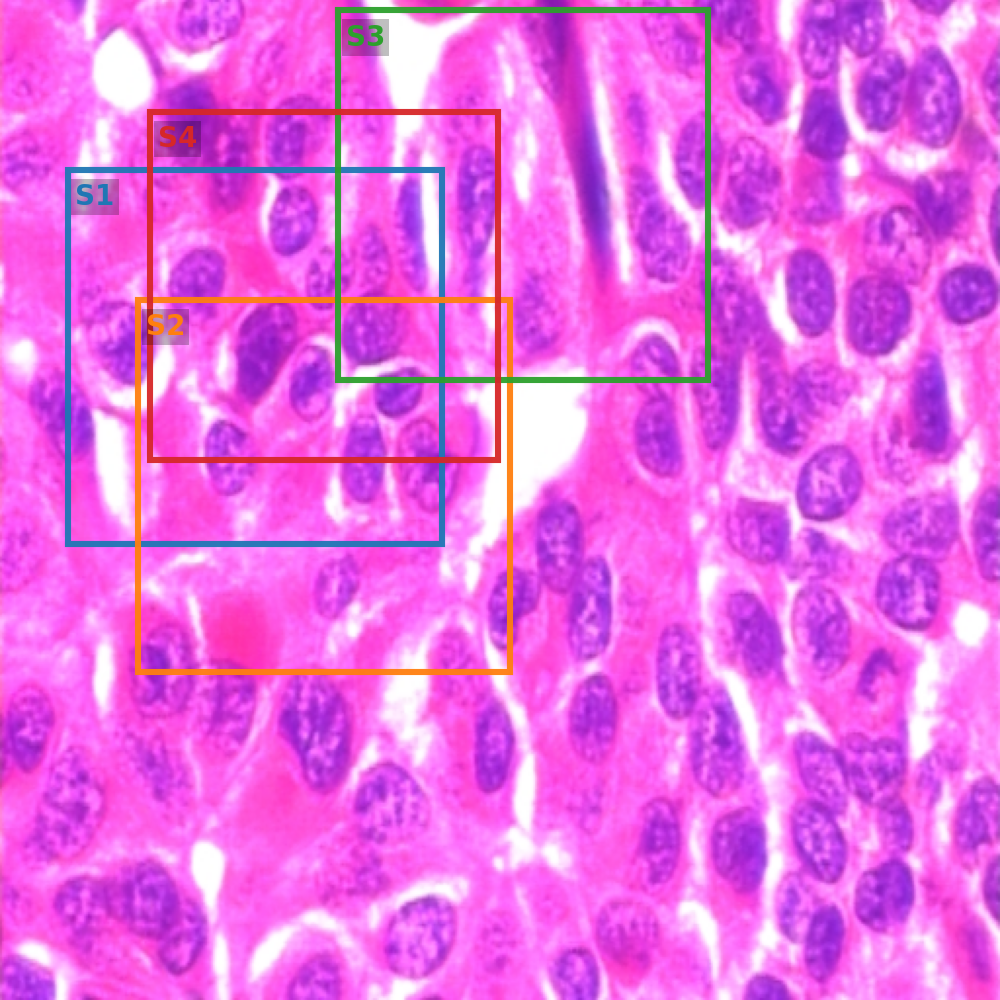}} &
\adjustbox{max width=\linewidth, max height=2.7cm}{\includegraphics{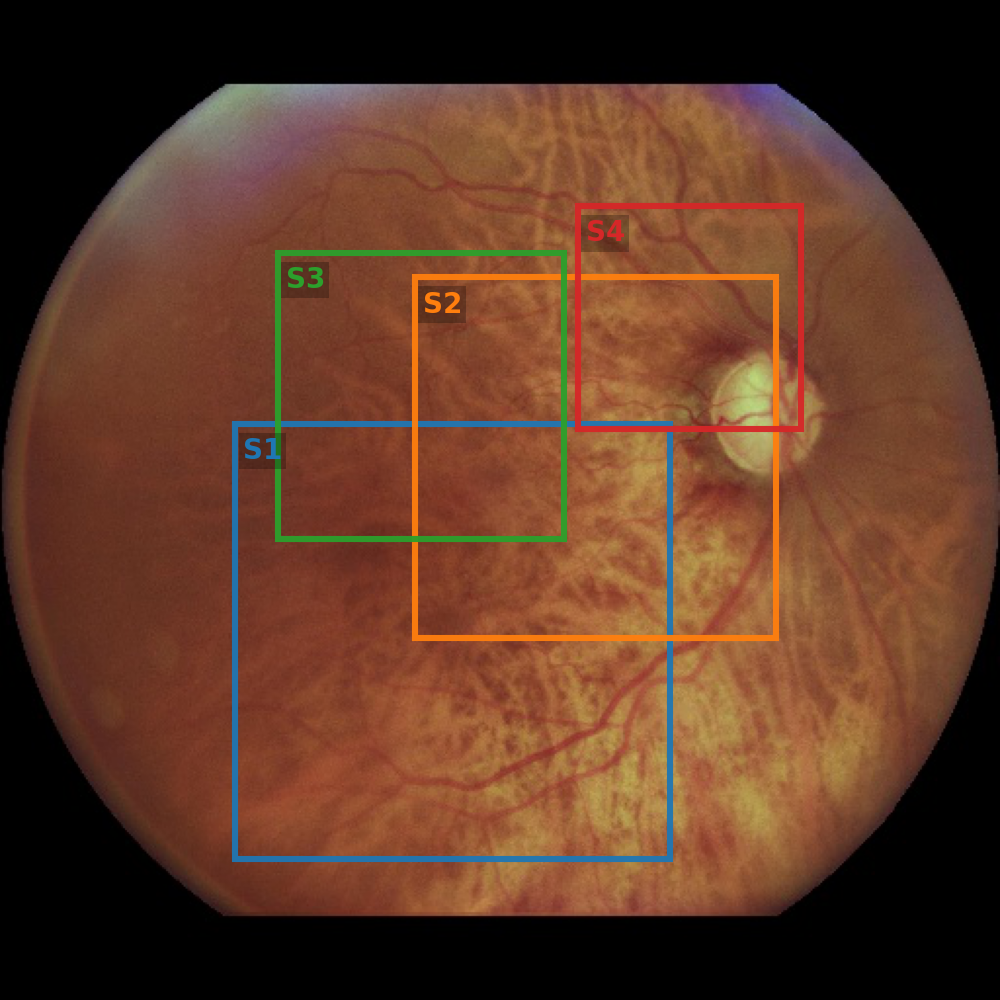}} \\
\end{tabular}

\caption{Progressive refinement boxes (Step1--Step4) on samples per dataset.}
\label{fig:visual}
\end{figure*}

\noindent\textbf{Efficiency.}As illustrated in Figure~\ref{fig:tradeoff_covid_busi}, \texttt{CPR} significantly shifts the accuracy and efficiency trade-off by decoupling high-resolution performance from computational cost. By selectively refining salient regions rather than processing full images uniformly, \texttt{CPR} achieves a $19.6\times$ reduction in GFLOPs on the COVID-19 dataset with a ResNet-50 backbone while maintaining comparable accuracy to high-resolution baselines. This efficiency extends across architectures; for the ViT-S backbone, \texttt{CPR} yields a $6.3\times$ and $3.5\times$ FLOP reduction on COVID-19 and BUSI, respectively. Furthermore, as shown in Figure~\ref{fig:fig1}, \texttt{CPR} maintains a constant GPU memory footprint regardless of input resolution, effectively bypassing the quadratic memory growth and Out-of-Memory (OOM) constraints that limit standard models like ViT-S at resolutions beyond $384\times384$.


\begin{table}[tbh]
\centering
\caption{Ablation study on scale control designs.
Base denotes the fine-tuning from pre-trained ViT-S; 
Base+$(x_t, y_t)$ uses fixed scale ($s_t$=1); 
Base+$(x_t, y_t, s_t)$ predicts scale adaptively; 
Base+$(x_t, y_t, s_t)+\phi_t^{\text{scale}}$ adds explicit scale control; 
\texttt{CPR} denotes Base+$(x_t, y_t, s_t)+\phi_t^{\text{scale}}+\phi_t^{\text{rep}}$.}
\label{tab:scale_ablation}
\resizebox{0.63\linewidth}{!}{
\begin{tabular}{
l
@{\hspace{12pt}}c@{\hspace{18pt}}   
@{\hspace{6pt}}c@{\hspace{10pt}}   
c c c
}
\toprule
Dataset & Base & \makecell{Base+\\$(x_t,y_t)$} &
\makecell{Base+\\$(x_t,y_t,s_t)$} &
\makecell{Base+\\$(x_t,y_t,s_t)+\phi_t^{scale}$} &
\cellcolor{cprred}\texttt{CPR} \\
\midrule
Shenzhen & 86.36 & 87.12 & 87.12 & 89.39 & \cellcolor{cprred}\textbf{90.15} \\
COVID-19 & 96.35 & 96.98 & 97.52 & 98.19 & \cellcolor{cprred}\textbf{98.67} \\
BUSI & 92.70 & 93.33 & 93.33 & 93.65 & \cellcolor{cprred}\textbf{94.89} \\
BreaKHis & 97.25 & 97.98 & 98.35 & 98.53 & \cellcolor{cprred}\textbf{99.27} \\
EyePACS & 91.95 & 92.47 & 93.25 & 94.94 & \cellcolor{cprred}\textbf{95.19} \\
\bottomrule
\end{tabular}}
\end{table}
\subsection{Additional Analysis}
\begin{figure*}[tbh]
\centering
\includegraphics[width=0.9\textwidth]{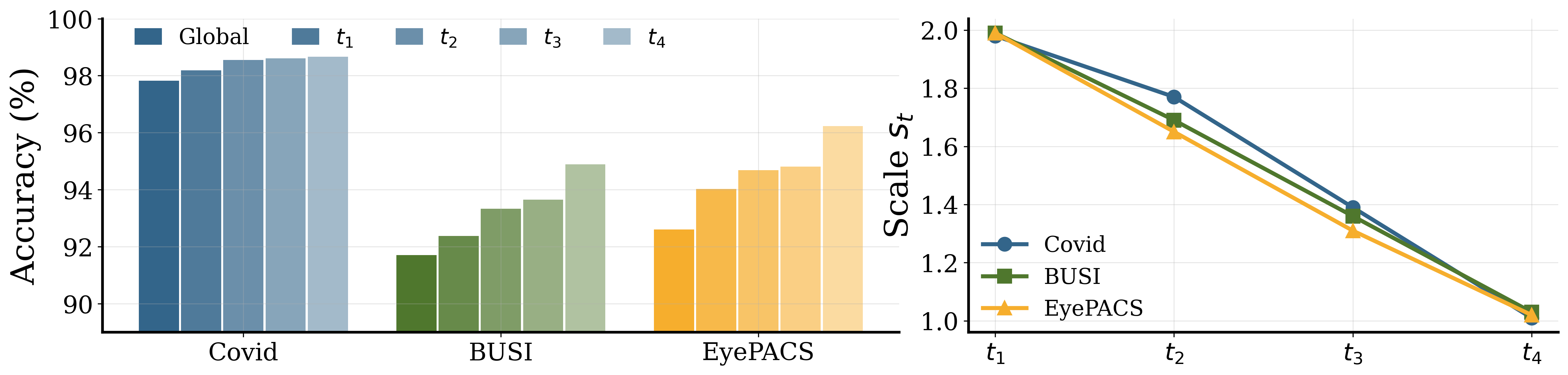}\caption{Qualitative visualization of progressive refinement. (Left:) Accuracy from global to progressive local refinement. (Right:) Scale trend under refinement steps.}
\label{fig:cpr_qual_abc}
\end{figure*}

\noindent\textbf{Visualization.}
Fig.~\ref{fig:visual} shows that \texttt{CPR} progressively refines perceptual fields across datasets, moving from global context to discriminative regions. 
This coarse-to-fine process reduces redundant computation and improves spatial efficiency. 
The refinement boxes consistently contract toward clinically relevant structures, indicating stable and meaningful scale adaptation.

\noindent\textbf{Ablations.}
Table~\ref{tab:scale_ablation} compares different scale control formulations.
Using only $(x_t,y_t)$ already improves over the backbone but remains inferior, indicating that spatial localization alone is insufficient.
Predicting $(x_t,y_t,s_t)$ yields modest gains on some datasets yet is unstable (e.g., Shenzhen shows no improvement over $(x_t,y_t)$).
Introducing the explicit scale-control term $(x_t,y_t,s_t)+\phi_t^{\text{scale}}$ consistently boosts performance.
The full \texttt{CPR} model achieves the best results across all datasets, validating structured scale regulation and coordinated refinement.

Figure~\ref{fig:cpr_qual_abc} further illustrates progressive refinement in terms of accuracy and scale. As shown in (a), accuracy steadily increases from the global stage to later refinement steps on COVID-19, BUSI, and EyePACS, confirming that each stage contributes complementary discriminative evidence. In (b), the predicted scale monotonically decreases across steps, reflecting the intended coarse-to-fine mechanism. Together, these findings verify that \texttt{CPR} concentrates computation on informative regions while consistently improving predictive accuracy.

\begin{table}[tbh]
\centering
\scriptsize

\begin{minipage}[t]{0.43\linewidth}
\centering
\caption{Additional control variants on BUSI using ViT-S.}
\label{tab:control_variants}
\setlength{\tabcolsep}{3pt}
\renewcommand{\arraystretch}{0.86}
\begin{tabular}{lc}
\toprule
Method & Acc. (\%) \\
\midrule
Base & 92.70 \\
Random Crop & 93.29 \\
Fixed Coarse-to-Fine & 93.61 \\
Deterministic Attention & 94.25 \\
No-Chain CPR & 94.25 \\
\rowcolor{cprred} \texttt{CPR} & \textbf{94.89} \\
\bottomrule
\end{tabular}
\end{minipage}
\hfill
\begin{minipage}[t]{0.55\linewidth}
\centering
\caption{Comparison with related adaptive architectures. (\%)}
\label{tab:related_adaptive}
\setlength{\tabcolsep}{2.5pt}
\renewcommand{\arraystretch}{0.86}
\begin{tabular}{lcc}
\toprule
Method & BUSI & COVID-19 \\
\midrule
RA-CNN~\cite{fu2017racnn} & 88.82 & 98.25 \\
SSPS/M2Former~\cite{moon2023m2former} & 93.61 & 95.71 \\
\makecell[l]{Focus Longer\\to See Better~\cite{shroff2020focus}} & 90.10 & 95.59 \\
TNet~\cite{papadopoulos2021tnet} & 94.25 & 98.37 \\
\rowcolor{cprred} \texttt{CPR} & \textbf{94.89} & \textbf{98.67} \\
\bottomrule
\end{tabular}
\end{minipage}

\end{table}

\noindent\textbf{Additional controls, baselines, and stability.}
Table~\ref{tab:control_variants} shows that \texttt{CPR} outperforms random cropping, fixed coarse-to-fine scheduling, deterministic attention, and No-Chain CPR, confirming that the improvement benefits from both adaptive policy learning and chained state accumulation. Table~\ref{tab:related_adaptive} further shows that \texttt{CPR} surpasses related iterative/adaptive architectures on BUSI and COVID-19. Separately, to assess policy-learning stability, we repeat \texttt{CPR} with three random seeds and obtain $98.64\pm0.17\%$ on COVID-19 and $95.16\pm0.30\%$ on BUSI, indicating stable training. Failure cases mainly involve diffuse or weakly contrasted lesions, where over-contracted crops may discard useful context, or salient non-lesion structures that may mislead the policy.

\section{Conclusion}


We present Chained Perceptual Refinement (\texttt{CPR}), a resolution-aware framework that reformulates high-resolution medical image analysis as a structured, coarse-to-fine reasoning process. By dynamically attending to informative regions via an adaptive policy, CPR sidesteps the prohibitive memory growth typically associated with high-resolution architectures. Our evaluation across five diverse public benchmarks demonstrates that CPR consistently surpasses both fixed-resolution baselines and state-of-the-art multi-scale methods, all while maintaining a constant, hardware-efficient memory footprint.

\noindent\textbf{Acknowledgments.}
This work was partially supported by the National Natural Science Foundation of China (Grant No. 62501295), the Natural Science Research Start-up Foundation of Recruiting Talents of Nanjing University of Posts and Telecommunications (Grant No. XK0020923211), and the Jiangsu Provincial Distinguished Postdoctoral Program (2024ZB490).

\noindent\textbf{Disclosure of Interests.}
The authors have no competing interests to declare that are relevant to the content of this paper.
\bibliographystyle{splncs04}
\bibliography{references}

@inproceedings{fu2017racnn,
  title     = {{Look Closer to See Better}: Recurrent Attention Convolutional Neural Network for Fine-Grained Image Recognition},
  author    = {Fu, Jianlong and Zheng, Heliang and Mei, Tao},
  booktitle = {Proceedings of the IEEE Conference on CVPR},
  pages     = {4438--4446},
  year      = {2017}
}

@article{moon2023m2former,
  title   = {{M2Former}: Multi-Scale Patch Selection for Fine-Grained Visual Recognition},
  author  = {Moon, Jiyong and Lee, Junseok and Lee, Yunju and others},
  journal = {arXiv preprint arXiv:2308.02161},
  year    = {2023}
}

@inproceedings{shroff2020focus,
  title     = {{Focus Longer to See Better}: Recursively Refined Attention for Fine-Grained Image Classification},
  author    = {Shroff, Prateek and Chen, Tianlong and Wei, Yunchao and others},
  booktitle = {CVPR},
  pages     = {960--961},
  year      = {2020}
}

@inproceedings{papadopoulos2021tnet,
  title     = {Hard-Attention for Scalable Image Classification},
  author    = {Papadopoulos, Athanasios and Korus, Pawe{\l} and Memon, Nasir},
  booktitle = {Advances in NIPS},
  volume    = {34},
  pages     = {14694--14707},
  year      = {2021}
}

@inproceedings{pamil2024,
  title     = {Dynamic Policy-Driven Adaptive Multi-Instance Learning for Whole Slide Image Classification},
  author    = {Zheng, Tingting and Jiang, Kui and Yao, Hongxun},
  booktitle = {CVPR},
  year      = {2024}
}

@article{gbmil2025,
  title   = {A Spatially-Aware Multiple Instance Learning Framework for Digital Pathology},
  author  = {Keshvarikhojasteh, Hassan and Tifrea, Mihail and others},
  journal = {arXiv preprint arXiv:2504.17379},
  year    = {2025}
}

@inproceedings{swinvit,
  title     = {Swin Transformer: Hierarchical Vision Transformer Using Shifted Windows},
  author    = {Liu, Ze and Lin, Yutong and Cao, Yue and others},
  booktitle = {ICCV},
  pages     = {10012--10022},
  year      = {2021}
}

@article{zhu2024glnet,
  title   = {Revisiting the Integration of Convolution and Attention for Vision Backbone},
  author  = {Zhu, Lei and Wang, Xinjiang and Zhang, Wayne and others},
  journal = {Advances in NIPS},
  year    = {2024}
}

@article{ZoomNeXt,
  title   = {ZoomNeXt: A Unified Collaborative Pyramid Network for Camouflaged Object Detection},
  author  = {Pang, Youwei and Zhao, Xiaoqi and Xiang, Tian-Zhu and others},
  journal = {IEEE TPAMI},
  year    = {2024}
}

@article{ryali2023hiera,
  title   = {Hiera: A Hierarchical Vision Transformer without the Bells-and-Whistles},
  author  = {Ryali, Chaitanya and Hu, Yuan-Ting and Bolya, Daniel and others},
  journal = {ICML},
  year    = {2023}
}

@article{mgca,
  title     = {Tuberculosis and pneumonia diagnosis in chest X-rays by large adaptive filter and aligning normalized network with report-guided multi-level alignment},
  author    = {Lu, Si-Yuan and Zhu, Ziquan and Zhang, Yu-Dong and others},
  journal   = {Engineering Applications of Artificial Intelligence},
  volume    = {158},
  pages     = {111575},
  year      = {2025},
  publisher = {Elsevier}
}

@inproceedings{eyepac,
  title        = {Automated fundus image standardization using a dynamic global foreground threshold algorithm},
  author       = {Kiefer, Riley and Abid, Muhammad and Ardali, Mahsa Raeisi and others},
  booktitle    = {International Conference on Image, Vision and Computing},
  pages        = {460--465},
  year         = {2023},
  organization = {IEEE}
}

@article{breakhis,
  title   = {A Dataset for Breast Cancer Histopathological Image Classification},
  author  = {Spanhol, Fabio A. and Oliveira, Luiz S. and Petitjean, Caroline and others},
  journal = {IEEE Transactions on Biomedical Engineering},
  year    = {2016},
  volume  = {63},
  number  = {7},
  pages   = {1455--1462}
}

@inproceedings{mammo-clip,
  title     = {Mammo-CLIP: A Vision Language Foundation Model to Enhance Data Efficiency and Robustness in Mammography},
  author    = {Ghosh, Shantanu and Poynton, Clare B. and Visweswaran, Shyam and others},
  booktitle = {MICCAI 2024},
  year      = {2024},
  publisher = {Springer Nature Switzerland},
  pages     = {632--642}
}

@article{wang2025emulating,
  title     = {Emulating human-like adaptive vision for efficient and flexible machine visual perception},
  author    = {Wang, Yulin and Yue, Yang and others},
  journal   = {Nature Machine Intelligence},
  pages     = {1--19},
  year      = {2025},
  publisher = {Nature Publishing Group UK London}
}

@article{meng_lou_overlock_2025,
	title = {{OverLoCK}: {An} {Overview}-first-{Look}-{Closely}-next {ConvNet} with {Context}-{Mixing} {Dynamic} {Kernels}},
	journal = {arXiv preprint arXiv:2502.20087},
	author = {Meng Lou and Yizhou Yu},
	year = {2025},
	pages = {1--15},
}

@article{jaeger_s_automatic_2014,
	title = {Automatic tuberculosis screening using chest radiographs},
	volume = {33},
	number = {2},
	journal = {IEEE Trans Med Imaging},
	author = {Jaeger S, Karargyris A.},
	year = {2014},
	pages = {233--245},
}

@misc{alexey_dosovitskiy_image_2021,
	address = {Virtual},
	title = {An {Image} {Is} {Worth} 16x16 {Words}: {Transformers} for {Image} {Recognition} at {Scale}},
	author = {Alexey Dosovitskiy and Lucas Beyer and Alexander Kolesnikov and Dirk Weissenborn and others},
	year = {2021},
}

@article{murugesan_neural_2024,
	chapter = {5043},
	title = {Neural {Networks} {Based} {Smart} {E}-{Health} {Application} for the {Prediction} of {Tuberculosis} {Using} {Serverless} {Computing}},
	volume = {28},
	issn = {2168-2194 2168-2208},
	number = {9},
	journal = {‌IEEE Journal of Biomedical and Health Informatics‌},
	author = {Murugesan, Subramaniam Subramanian and Velu, Sasidharan and others},
	year = {2024},
	pages = {5043--5054},
}

@article{vats_incremental_2024,
	chapter = {122129},
	title = {Incremental learning-based cascaded model for detection and localization of tuberculosis from chest x-ray images},
	volume = {238},
	issn = {09574174},
	journal = {Expert Systems with Applications},
	author = {Vats, Satvik and Sharma, Vikrant and Singh, Karan and others},
	year = {2024},
}

@article{serrao_automatic_2024,
	chapter = {108167},
	title = {Automatic bright-field smear microscopy for diagnosis of pulmonary tuberculosis},
	volume = {172},
	issn = {00104825},
	journal = {Comput Biol Med},
	author = {Serrão, Mikaela Kalline Maciel and Costa, Marly Guimarães Fernandes and Fujimoto, Luciana Botinelly Mendonça and Ogusku, Mauricio Morishi and Costa Filho, Cicero Ferreira Fernandes},
	year = {2024},
}

@article{wang_joint_2024,
	chapter = {103032},
	title = {Joint learning framework of cross-modal synthesis and diagnosis for {Alzheimer}’s disease by mining underlying shared modality information},
	volume = {91},
	issn = {13618415},
	journal = {Med Image Anal},
	author = {Wang, Chenhui and Piao, Sirong and Huang, Zhizhong and Gao, Qi and Zhang, Junping and Li, Yuxin and Shan, Hongming},
	year = {2024},
}

@article{yu_siamese-transport_2024,
	chapter = {391},
	title = {A {Siamese}-{Transport} {Domain} {Adaptation} {Framework} for {3D} {MRI} {Classification} of {Gliomas} and {Alzheimer}'s {Diseases}},
	volume = {28},
	issn = {2168-2194 2168-2208},
	number = {1},
	journal = {‌IEEE Journal of Biomedical and Health Informatics‌},
	author = {Yu, Luyue and Liu, Ju and Wu, Qiang and Wang, Jing and Qu, Aixi},
	year = {2024},
	pages = {391--402},
}

@article{al-dhabyani_dataset_2020,
	title = {Dataset of breast ultrasound images},
	volume = {28},
	issn = {2352-3409 (Electronic) 2352-3409 (Linking)},
	journal = {Data Brief},
	author = {Al-Dhabyani, W. and Gomaa, M. and Khaled, H. and Fahmy, A.},
	month = feb,
	year = {2020},
	pages = {104863},
}

@misc{kaiming_he_deep_2016,
	address = {Las Vegas, NV, USA},
	title = {Deep {Residual} {Learning} for {Image} {Recognition}},
	author = {Kaiming He and Xiangyu Zhang and Shaoqing Ren and Jian Sun},
    booktitle = {CVPR},
    year = {2016},
}

@article{muhammad_e_h_chowdhury_can_2020,
	title = {Can {AI} help in screening {Viral} and {COVID}-19 pneumonia?},
	journal = {IEEE Access},
	author = {Muhammad E. H. Chowdhury and Tawsifur Rahman and Amith Khandakar and Rashid Mazhar and Muhammad Abdul Kadir and Zaid Bin Mahbub and Khandaker Reajul Islam and Muhammad Salman Khan and Atif Iqbal and Nasser Al-Emadi and Mamun Bin Ibne Reaz},
	year = {2020},
    volume = {8},
    pages ={132665-132676},
}

@article{ashman2022whole,
  title={Whole slide image data utilization informed by digital diagnosis patterns},
  author={Ashman, Kimberly and Zhuge, Huimin and Shanley, Erin and Fox, Sharon and Halat, Shams and Sholl, Andrew and Summa, Brian and Brown, J Quincy},
  journal={Journal of Pathology Informatics},
  volume={13},
  pages={100113},
  year={2022},
  publisher={Elsevier}
}

@article{sabottke2020effect,
  title={The effect of image resolution on deep learning in radiography},
  author={Sabottke, Carl F and Spieler, Bradley M},
  journal={Radiology: Artificial Intelligence},
  volume={2},
  number={1},
  pages={e190015},
  year={2020},
  publisher={Radiological Society of North America}
}

@article{haque2023effect,
  title={Effect of image resolution on automated classification of chest X-rays},
  author={Haque, Md Inzamam Ul and Dubey, Abhishek K and Danciu, Ioana and Justice, Amy C and Ovchinnikova, Olga S and Hinkle, Jacob D},
  journal={Journal of Medical Imaging},
  volume={10},
  number={4},
  pages={044503--044503},
  year={2023},
  publisher={Society of Photo-Optical Instrumentation Engineers}
}

\end{document}